# Learning Unit State Recognition Based on Multi-channel Data Fusion

Feng Tian
National Engineering Lab of Big Data Analytics
*Xi'An Jiaotong University*
Xi'an, CHINA
fengtian@mail.xjtu.edu.cn

Jia Yue, Xing Wan
MOE KLINNS
*Xi'An Jiaotong University*
Xi'an, CHINA
{ marshall, wanxing}@stu.xjtu.edu.cn

Kuo-Min Chao
School of Computing, Electronics and Mathematics,
Coventry University
Coventry, UK
csx240@coventry.ac.uk

Qinghua Zheng
MOE KLINNS
*Xi'An Jiaotong University*
Xi'an, CHINA
qhzheng@mail.xjtu.edu.cn

**Abstract**

*Despite recent advances in MOOC, the current e-learning systems have advantages of alleviating barriers by time differences, and geographically spatial separation between teachers and students. However, there has been a 'lack of supervision' problem that e-learner's learning unit state(LUS) can't be supervised automatically. In this paper, we present a fusion framework considering three channel data sources: 1) videos/images from a camera, 2) eye movement information tracked by a low solution eye tracker and 3) mouse movement. Based on these data modalities, we propose a novel approach of multi-channel data fusion to explore the learning unit state recognition. We also propose a method to build a learning state recognition model to avoid manually labeling image data. The experiments were carried on our designed online learning prototype system, and we choose CART, Random Forest and GBDT regression model to predict e-learner's learning state. The results show that multi-channel data fusion model have a better recognition performance in comparison with single channel model. In addition, a best recognition performance can be reached when image, eye movement and mouse movement features are fused.*

## 1. Introduction

The e-learning system has gradually become a trend for online education and learning in recent years. This new educational model[1] overcomes the limitations of space and time exiting in the traditional mode of education. At the same time, it provides abundant resources and reduces the cost of learning, which adds new connotation to the process of teaching and learning[2]. However, due to spatial and temporal separation between teachers and students, it is difficult for teachers to supervise the students' learning task. Unable to interact with each other may result in low learning efficiency[3]. To solve the 'lack of supervision' problem[4][5][6], it is imperative to recognize the learning state of students.

The learner's learning state is a result of the combination of mind, body condition and learning methods, usually expressed through face expression, body gesture, speech, and physiological signal. An integrated consideration of emotion and behavior using single or multi-channel data sources exists in current researches, which have shown considerable promise to improve learning efficiency. Asteriadis al.[7] estimated the learning state of user based on eye gaze and head pose from video channel in an e-learning environment, Azcarraga al.[8] predicted academic emotion by using brainwaves signals and mouse movement combined with personality profile, Klašnja-Milićević al.[9] adopted an eye-tracking technique to analyze the e-learners' cognitive procedure, Feng al.[10] recognized and regulated e-learners' emotion in interactive Chinese texts, S. Saha proposed a system to classify learning state based on body gesture[11]. Nevertheless, research[12] suggests that fusing the multimodal data result in a large increase in the recognition rates in comparison with the unimodal systems. Soujanya Poria al.[13] demonstrated a model that fuses audio, visual and textual modalities for real-time sentiment analysis, which needs quite a bit annotations. Luc Paquette al.[14] built multi-channel affect detection models based on two different data modalities: software logs and posture data.

However, the researches mentioned above mostly recognized emotions and behaviors of the instantaneous learning state in real time which require manually labelling learning states. Synchronous multiple data channels increase the difficulty of labelling, so there is no relevant open dataset available. We overcome the above issue by adopting a small period of time instead of real time to recognize learning state. A period of time, from 5 to 15 minutes, is measured by a learning topic, or a short video clip. A collection of a learner's learning states over a period of time is learning unit state. In order to solve the problem of labeling data manually, this paper proposes a learning unit state labeling method that combines various related factors such as e-learner's foundational capabilities, self-evaluation and class-evaluation, which automatically forms a LUS dataset.

Aiming at the 'lack of supervision' problem, this paper proposes a framework that fuses multi-channel data including face expression, head posture, eye movement information and mouse movements to recognize the learning unit state. Firstly, a fusion framework combined three data modalities at the feature-level. In video channel, both spatial and temporal characteristics of video image sequence are extracted by a convolutional neural network(CNN)[15] conjunction with a long short-term memory(LSTM)[16] model. A migration learning method is adopted to train convolution layer parameters for emotion recognition based on image squeneces. Simultaneously, the features of eye and mouse movements in time sequences are extracted via statistical and wavelet methods. Then, these features from different data channels are integrated after filtering and selecting. Last but not least, a prototype system is designed to validate our proposed method, which provides technical support to overcome the problem of 'lack of supervision' in e-learning.

## 2. Multi-channel Data Fusion Framework

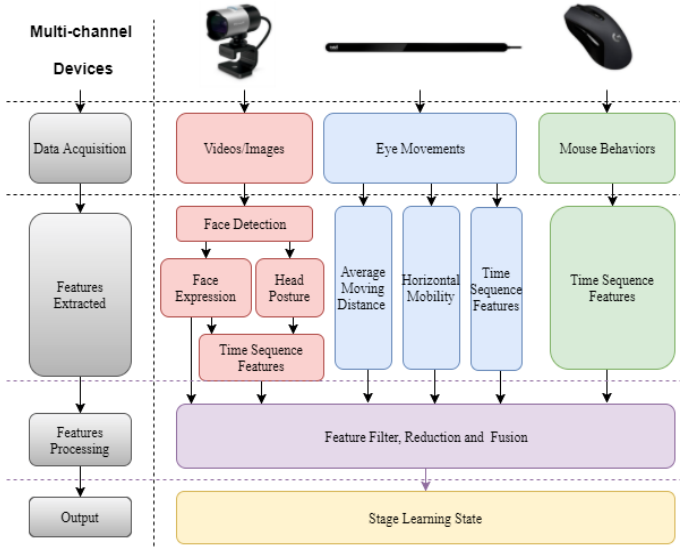

Figure 1. Multi-channel Data Fusion Framework

We choose three data acquisition devices: a low-cost camera for video recording, a low resolution eye tracker and a mouse. But we need to address two challenges: dimensional inconsistence in data channels, and alignment different frequency data. These can be addressed in feature- or decision-level fusion. The decision-level fusion leads to precision decline, so we choose a feature-level fusion to recognize e-learner's learning unit state. Figure 1 shows its framework.

In this framework, some core features and modules are introduced: 1)face features extraction and emotion recognition, 2) head posture features extraction, 3)eye tracker information and mouse movement features extraction and 4)feature selection, reduction and fusion.

### 2.1. Face Expression

According to the intensity and order of emotions sequence: onset, apex and offset[17], we will take face image sequence as an input of models instead of static image. The spatial-temporal features can be extracted by two steps:1) spatial image characteristics of the representative expression-state frames are learned via a CNN, 2) temporal characteristics of the spatial feature of the facial expression are learned with a LSTM. However, an intricate problem is lack of free eastern facial database leading to not enough numbers of training data for generalization of the trained models. Therefore, this paper adopts a migration learning method to learn two pre-training convolutional neural networks: VGG16[18] and Inception-ReNet-V2[19]. These two networks are trained with the ImageNet database which has more than 14 million images, to determine their pre-trained parameters. Then, these parameters can be fine-tuned by further training on the database to finalize the CNN models. Next, the spatial feature parameters of CNN models are fed to LSTM to train the temporal feature parameters. Furthermore, we employ USTC-NVIE database[20] from University of Science and Technology of China to learn Asia face features, because our experimental subjects are predominantly Chinese students.

To select an optimal model from the following four combinations: VGG16 without LSTM, Inception-ResNetV2 without LSTM, VGG16 with LSTM, and Inception-ResNetV2 with LSTM, we apply a 10-fold cross validation method to these four combinations and their precision rates are 71.54%, 72.32%, 76.08% and 75.47% respectively. Therefore, we select VGG16 with LSTM model with a highest precision rate.

### 2.2. Head Posture

Heading posture is a set of direction parameters in space coordinate system and they are horizontal rotation angle of yaw, vertical rotation of pitching angle, and rotation angle of rolling. The head posture features will be normalized to [-1,1].

### 2.3. Eye Tracking and Mouse Movement

Two subjective features and more than forty general features are extracted from eye movement information. The first subjectieve feature is derived from average moving distance between gaze points and the second one is horizontal mobility. The general features are calculated in a period of time, including statistical, wavelet and Fourier transform features.

Mouse movement shares the same general features with eye movement, so they also have similar methods to calculate

### 2.4. Feature Filter, Reduction and Fusion

As can be seen in Figure 2, after extracting the three data channel features over a period of time, the obtained features are filtered by a hypothesis test within the respective channel. Next, PCA dimension reduction method is adopted to squeeze a high features dimension to a low one. Then, in order to avoid the data alignment with different frequencies, we extract the features in a fixed time interval, before fusing multi-channel data features. Further, analyze the correlation of the combined features.

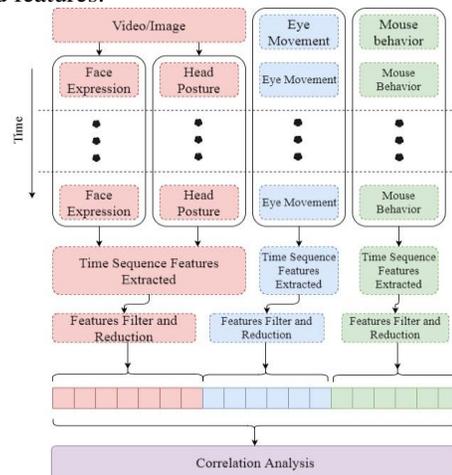

Figure 2. Features Filter, Reduction and Fusion

## 3. Experiment Results

## 3.1. Multi-channel Database

The course to learn is '*Data Processing Using Python*' on Coursera. We choose the first chapter as learning unit for e-learners, the unit contains three parts: video, lectures and evaluation. The steps of experiment in our online learning environment are: 1) e-learners are required to fill in information including name, major, sex, age and mastery degree of knowledge, 2) starting with learning the short videos one by one, e-learners are asked to self-evaluate from 10 to 100 based on the performance after learning each video, 3) complete the class-evaluation independently, after finished learning content.

The specific multi-channel data acquisition devices we chose are: Microsoft LifeCam camera, Tobii Eye Tracker 4C, and Logistic G603 mouse. During the learning process, the camera records a video with 15FPS and 640*480 resolution, eye track records e-learner's fixations with <*timestamp, event_type, x, y*> format, and mouse collector records the trail with <*message, time, x, y, wheel*> format. We collected 15 e-learners' multi-channel data and each e-learner spends approximate one hour to complete the learning task and exercise. The size of collected database is 19GB including e-learner's profile and self-evaluation data.

## 3.2. Results

1）Correlation analysis of multi-channel features

After dimension reduction through PCA with 95% retention of information, the ultimate number of dimension is shown in Table 2.

Table 1. Multi-channel data retention dimensions

| Channel | Retention dimensions |
|---|---|
| Video(face expression and head posture) | 249 |
| Eye movement | 44 |
| Mouse behavior | 285 |

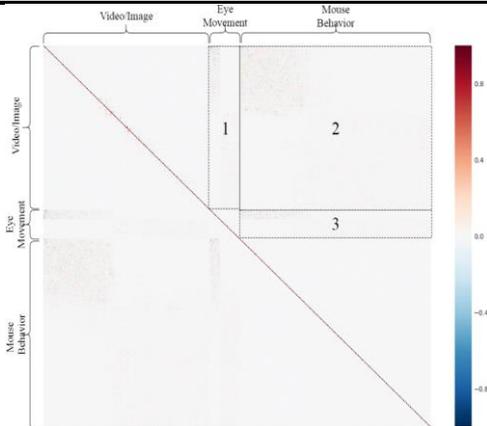

Figure 3 Correlation Analysis of Channel Features

Figure 3 illustrates the correlation of features of three data channels, which indicates an extremely weak correlation among these features. Therefore, the three channel features can be considered to be independent of each other, i.e. these features are able to provide complementary information to raise the robustness of the following regression models.

2）Regression Model Results Comparison

We treat the learning state recognition as regression problems instead of classification problems. We adopt the 10-fold cross validation method to train CART[21], Random Forest[22] and GBDT[23] regression models on the feature training sets which are a combination of different channel features, and then compare the learning unit state performance of three regression models in these features with the evaluation indicator $R^2$.

Table 2. Regression models performance comparison

| Channel or combined channels | CART | Random Forest | GBDT |
|---|---|---|---|
| Image | 0.839 | 0.920 | 0.923 |
| Eye movement | 0.306 | 0.425 | 0.439 |
| Mouse behavior | 0.829 | 0.917 | 0.915 |
| Image + Eye movement | 0.877 | 0.959 | 0.964 |
| Eye movement+ Mouse behavior | 0.869 | 0.951 | 0.959 |
| Image + Mouse behavior | 0.889 | 0.968 | 0.961 |
| Image +Eye movement + Mouse behavior | **0.905** | **0.979** | **0.980** |

Table 2 summarizes the results of LUS recognition, from which it can be seen that performance of fusion features models outperforms single channel features models. It is obvious that LUS fusing video, eye movement and mouse movement features reaches a best performance. In terms of comparison on the regression models' results, only a small difference in performance between Random Forest and GBDT, which have better performance than CATR.

## 4. A prototype system

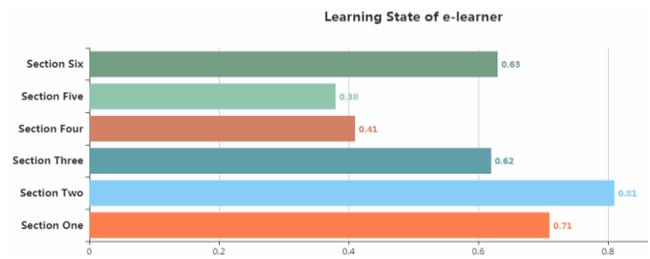

Figure 4. Learning State Recognition

A prototype system was designed with a Browser Server and Client Server framework. The multi-channel data are collected and processed by this system during learning course, and the results can be visually displayed on the web page. For example, Figure 4 shows an e-learner's LUS prediction results, and Figure 5 illustrates a real time face expression.

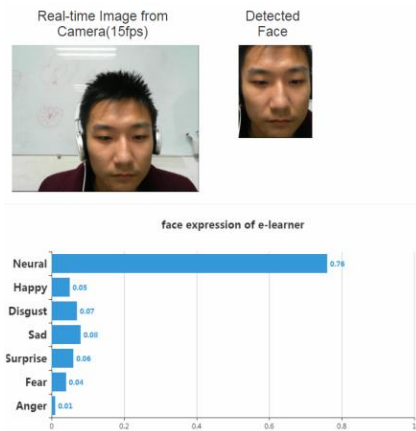

Figure 5. Real time face expression

## 5. Conclusion and Future work

In this paper, we have presented an multi-channel data fusion framework to recognize the learning unit state of e-learner, to address the "lack of supervision" issue. Moreover, we propose an approach that combines e-learner's foundational capability, self-evaluation and class-evalution to build a learning state dataset, to avoid labeling real time learning state manually. Furthermore, we fuse multiple data features on time sequence to solve the "time alignment". Additionally, three multi-channels data are collected by relatively low cost devices, and we obtain good results.

In the future work, we plan to expand our multi-channel dataset by inviting more people, and we will improve generalization of face expression recognition models through employing GAN (Generative Adversarial Networks) to augment database with limited samples. Meanwhile, to further improve the accuracy of our system, we will propose different fusion methods. Finally, we will devote more efforts into the research of e-learner's cognition.

## 6. Acknowledgment

This research was supported by the innovation team of the Ministry of Education IRT13035; the National Science Foundation of China under Grant No. 61402392, 61472315, 61428206, 61532015 and 61532004; The project of China Knowledge Centre for Engineering Science and Technology; The Natural Science Basic Research Plan in Shanxi Province of China under Grant No.2016JM6027, 2016JM6080.